\pdfoutput=1

\documentclass[11pt]{article}

\usepackage[final]{acl}

\usepackage{float}
\usepackage{times}
\usepackage{latexsym}
\usepackage{multirow}
\usepackage{graphicx}
\usepackage{adjustbox}
\usepackage{array}
\usepackage{enumitem} 

\usepackage[T1]{fontenc}

\usepackage[utf8]{inputenc}

\usepackage{microtype}

\usepackage{inconsolata}

\usepackage{graphicx}

\title{On Limitations of LLM as Annotator for Low Resource Languages}

\author{
    Suramya Jadhav\textsuperscript{1,3}, Abhay Shanbhag\textsuperscript{1,3}, Amogh Thakurdesai\textsuperscript{1,3}, 
    Ridhima Sinare\textsuperscript{1,3}, and Raviraj Joshi\textsuperscript{2,3} \\
    \textsuperscript{1}Pune Institute of Computer Technology, Pune \\
    \textsuperscript{2}Indian Institute of Technology Madras, Chennai \\
     \textsuperscript{3}L3Cube Labs, Pune
}

\begin{document}
\maketitle
\begin{abstract}

Low-resource languages face significant challenges due to the lack of sufficient linguistic data, resources, and tools for tasks such as supervised learning, annotation, and classification. This shortage hinders the development of accurate models and datasets, making it difficult to perform critical NLP tasks like sentiment analysis or hate speech detection. To bridge this gap, Large Language Models (LLMs) present an opportunity for potential annotators, capable of generating datasets and resources for these underrepresented languages. 
In this paper, we focus on Marathi, a low-resource language, and evaluate the performance of both closed-source and open-source LLMs as annotators, while also comparing these results with fine-tuned BERT models. We assess models such as GPT-4o and Gemini 1.0 Pro, Gemma 2 (2B and 9B), and Llama 3.1 (8B and 405B) on classification tasks including sentiment analysis, news classification, and hate speech detection. Our findings reveal that while LLMs excel in annotation tasks for high-resource languages like English, they still fall short when applied to Marathi. Even advanced models like GPT-4o and Llama 3.1 405B underperform compared to fine-tuned BERT-based baselines, with GPT-4o and Llama 3.1 405B trailing fine-tuned BERT by accuracy margins of 10.2\% and 14.1\%, respectively. This highlights the limitations of LLMs as annotators for low-resource languages.

\end{abstract}

\section{Introduction}



Even with advancements in NLP, the curation of annotations for supervised tasks like sentiment analysis, text classification, and inference has been the primary responsibility of human linguistic experts \cite{tan2024large}. Data annotations play an integral part in both building and evaluating a model. Hence, the quality and reliability of data lie at the core of the performance and usefulness of the model being built. 

The process of curating good-quality data annotations is expensive in terms of time and cost, specifically when it comes to compiling data annotations for low-resource languages. The aim of this study is to explore whether Large Language Models (LLMs) can be effectively leveraged to create supervised data resources for low-resource languages, with Marathi as the focus in this case.

Recent generative models like ChatGPT have shown competitive quality in data annotations for simpler tasks like sentiment analysis while human expert annotations proved to be better for intricate tasks \citet{Nasution2024ChatGPTLC}. 
ChatGPT was evaluated by \citet{zhu2023can} to check its capability of reproducing human-generated labels for social computing tasks. In these experiments, ChatGPT obtained an average accuracy of 0.60 with 0.64 being the highest accuracy for the sentiment analysis task. In addition to these, the works of \citet{kuzman2023chatgpt,gao2023human} have previously evaluated ChatGPT’s performance with that of human experts. Experiments performed by \citet{Mohta2023AreLL} demonstrated that Vicuna 13b performed reasonably well for numerous annotation tasks compared to other models that were tested like Vicuna 7b, Llama (13b, 7b) and InstructBLIP(13b, 7b). 
However, it is important to note that most of these experiments target the English language. 

India is a multilingual nation with various regional languages and most of these languages fall under the low-resource (LR) category. Low resource languages are languages such as Marathi and Hindi that lack annotated training datasets and have very few task-specific resources compared to high resource languages such as Spanish and English.

This paper presents a case study on the performance of Large Language Models (LLMs) in annotating the low-resource language Marathi. We conduct a comprehensive comparative analysis of various closed-source and open-source LLMs, revealing that many LLMs still fall significantly short of the baseline performance achieved by BERT-based models and are not yet capable of replacing human annotators.

Specifically, we evaluated models such as GPT-4o, Gemini 1.0 Pro, Gemma 2 (2B and 9B), Llama 3.1 (8B and 405B) across multiple tasks, including 3-class sentiment analysis, 2-class, and 4-class hate speech detection, as well as news classification based on headlines, long paragraphs, and full documents.

The key contributions of this research work are as follows:
\begin{itemize}[leftmargin=*, label=\textbullet]
    \item We have conducted a first-of-its-kind detailed comparative study between fine-tuned BERT models and large language models (LLMs), by evaluating their potential to be used as annotators for a low-resource language, Marathi.
    \item We observe that the average results of the Few-shot prompting technique outperform the average result of the Zero-shot prompting technique in all the models tested.
    \item We have provided valuable insights into the effectiveness of both open- and closed-source large language models (LLMs), including GPT-4o, Llama 3.1 405B, Llama 3.1 8B, Gemma 2 9B, Gemma 2 2B, and Gemini 1.0 Pro, on tasks such as Marathi Sentiment Analysis, Hate Speech Detection, and News Categorization. Our results strongly demonstrate that LLMs are still not fully reliable for annotation tasks in Indic languages.
    \item  Model ranking, based on
 accuracy metrics, is GPT-4o >
 Llama-3.1-405B > Gemini 1.0 Pro > Gemma 2 9B > Llama 3.1 8B > Gemma-2-2B.

\end{itemize}

The paper is structured as follows: Section 2 provides a concise review of prior research on data annotation and the use of LLMs. In Section 3, we detail the datasets used and the Section 4 describes models employed in our evaluation. Section 5 describes the experimental setup and the APIs leveraged to evaluate the LLMs. Section 6 presents the results, along with a comparative analysis of various LLMs and BERT-based models, highlighting the key findings of our research. Finally, in Section 7, we conclude our discussion.

\section{Literature Review}

Many low-resource languages, including Marathi, lack well-annotated datasets, making it difficult to train effective models for tasks like sentiment analysis and classification \citet{AlWesabi2023LowResourceLP}. The absence of sufficient data often leads to poor performance in tasks that require labeled corpora \citet{VR2023AnalysisOS}.

Low-resource languages also present unique linguistic challenges not well-represented in high-resource models \citet{Krasadakis2024ASO}, highlighting the need for specialized approaches. With the rise of LLMs, these models have been explored as a solution to mitigate the scarcity of annotated data in low-resource languages.

Several works demonstrate the use of LLMs as annotators for low-resource language tasks. \citet{Pavlovic2024TheEO} reviewed LLMs like GPT-4 and noted performance drops when handling non-English languages. 
In \citet{Kholodna2024LLMsIT}, the authors explored the integration of large language models (LLMs), specifically GPT-4 Turbo, into an active learning framework designed for low-resource language tasks. Their work demonstrates the use of few-shot learning to generate useful annotations, significantly enhancing performance on low-resource tasks.
Additionally, they implemented the GPT-4 Turbo model as a classifier within the training loop, leading to a substantial reduction in annotation costs, which were 42.45 times lower compared to traditional methods. However, the general performance of LLMs remains limited, especially for languages with fewer resources \citet{Hedderich2020ASO}.

The studies of \citet{Ding2022IsGA} and \citet{Mohta2023AreLL} further evaluated LLM performance on multilingual datasets, with results indicating that models like GPT-3 and open-source LLMs struggle with non-English data. \citet{srivastava2022beyond} showed that increasing model size does not consistently enhance performance for low-resource languages, unlike high-resource languages like English.

Bias is another concern with LLMs. \citet{Bavaresco2024LLMsIO} introduced JUDGE-BENCH to evaluate LLM biases, noting that training data heavily influences model outputs, which can be problematic in annotating complex or sensitive tasks in low-resource languages.
While LLMs used for high-resource languages (HRL) are giving promising results, that is not the case for low-resource languages.
 \citet{Nasution2024ChatGPTLC} explored ChatGPT-4’s performance in annotation tasks across languages like Turkish and Indonesian, offering insights into LLM applicability for Low Resource Language(LRL), a relevant consideration for our focus on Marathi.

\begin{table*}[t]
\centering
\resizebox{\textwidth}{!}{ 
\begin{tabular}{|c|c|c|c|c|c|c|c|c|}
\hline
\textbf{Dataset} & \textbf{Tech} & 
\textbf{Llama 3.1 8B}
 &
  \textbf{Gemma 2 2B}
  &
 \textbf{Gemma 2 9B} 
 & 
 \textbf{Gemini 1.0 Pro}
 & 
 \textbf{Llama 3.1 405B}  
 & 
 \textbf{GPT-4o}
 & 
 \textbf{Fine Tuned BERT} \\ \hline
 
\multirow{2}{*}{\href{https://github.com/l3cube-pune/MarathiNLP/tree/main/L3CubeMahaSent\%20Dataset}{\textbf{MahaSent}}} 
                           & ZS  & 0.76         & 0.71        & 0.69        & 0.78           & 0.77  & 0.79 & 0.80    \\  
                           & FS   & 0.79         & 0.76        & 0.78        & 0.76           & 0.81  & \textbf{0.82} &     \\ \hline
\multirow{2}{*}{\href{https://github.com/l3cube-pune/MarathiNLP/tree/main/L3Cube-MahaHate/2-class}{\textbf{MahaHate-2C}}} 
                           & ZS  & 0.64         & 0.71        & 0.78        & 0.74           & 0.77  & 0.80 & \textbf{0.91}    \\  
                           & FS   & 0.78         & 0.72        & 0.82        & 0.72           & 0.82  & 0.82 &    \\ \hline
\multirow{2}{*}{\href{https://github.com/l3cube-pune/MarathiNLP/tree/main/L3Cube-MahaHate/4-class}{\textbf{MahaHate-4C}}}
                           & ZS  & 0.40         & 0.39        & 0.43        & 0.43           & 0.49  & 0.58 & \textbf{0.73}    \\  
                           & FS   & 0.48         & 0.41        & 0.46        & 0.45           & 0.52  & 0.60 &     \\ \hline
\multirow{2}{*}{\href{https://github.com/l3cube-pune/MarathiNLP/tree/main/L3Cube-MahaNews/SHC}{\textbf{MahaNews-SHC}}}      
                           & ZS  & 0.60         & 0.54        & 0.68        & 0.68           & 0.75  & 0.78 & \textbf{0.85}    \\  
                           & FS   & 0.66         & 0.54        & 0.68        & 0.70           & 0.74  & 0.78 &     \\ \hline
\multirow{2}{*}{\href{https://github.com/l3cube-pune/MarathiNLP/tree/main/L3Cube-MahaNews/LPC}{\textbf{MahaNews-LPC}}}     
                           & ZS  & 0.66         & 0.55        & 0.71        & 0.72           & 0.76  & 0.77 & \textbf{0.89}    \\  
                           & FS   & 0.67         & 0.50        & 0.72        & 0.74           & 0.76  & 0.75 &     \\ \hline
\multirow{2}{*}{\href{https://github.com/l3cube-pune/MarathiNLP/tree/main/L3Cube-MahaNews/L}{\textbf{MahaNews-LDC}}}       
                           & ZS  & 0.69         & 0.62        & 0.78        & 0.74           & 0.76  & 0.81 & \textbf{0.96}    \\  
                           & FS   & 0.69         & 0.62        & 0.80        & 0.75           & 0.78  & 0.81 &    \\ \hline
\textbf{Average}           & ZS  & 0.625         & 0.587        & 0.678        & 0.682           & 0.716  & 0.755 & \textbf{0.857}  \\  
                           & FS  & 0.678         & 0.592        & 0.710        & 0.687           & 0.738  & 0.763 &  \\ \hline
\end{tabular}
}
\caption{Model Comparison across different tasks. Tech: Different Prompting Techniques Used; ZS: Zero Shot; FS: Few Shot; 2C: 2-Class; 4C: 4-Class; SHC: Short Headlines Classification; LDC: Long Document Classification; LPC: Long Paragraph Classification; BERT: Refer Section 4.2 for details about BERT models.}
\label{tab:Results}
\end{table*}

\section{Dataset}
 In this research, we focus on three major task categories using relevant Marathi datasets: 
 
 1) MahaSent \cite{kulkarni2021l3cubemahasent,pingle2023l3cube} – classifies sentiment of Marathi tweets into three classes of positive, negative, or neutral categories. 
 
 2) MahaHate \cite{patil2022l3cube} – measures the level of abusive and hostile content in Marathi text. This dataset includes two supervised tasks: MahaHate 2-Class, which categorizes content as either HATE or NOT, and MahaHate 4-Class, which provides finer distinctions with categories: Hate (HATE), Offensive (OFFN), Profane (PRFN), and Not (NOT). 
 
 3) MahaNews \cite{mittal2023l3cube,mirashi2024l3cube} – classifies headlines and articles from Marathi news. It comprises three supervised datasets: Short Headlines Classification (SHC), Long Document Classification (LDC), and Long Paragraph Classification (LPC), each categorizing news content into 12 classes: Auto, Bhakti, Crime, Education, Fashion, Health, International, Manoranjan, Politics, Sports, Tech, and Travel. The distribution of all the mentioned datasets is provided in Table 2.

 \begin{table}[h!]
    \centering
    \renewcommand{\arraystretch}{6} 
    \resizebox{\columnwidth}{!}{ 
    \begin{tabular}{|c|c|c|c|c|c|c|}
        \hline
        {\fontsize{50}{14}\selectfont Split} & 
        {\fontsize{50}{14}\selectfont MahaSent} & 
        {\fontsize{50}{14}\selectfont MahaHate 2-C} & 
        {\fontsize{50}{14}\selectfont MahaHate 4-C} & 
        {\fontsize{50}{14}\selectfont SHC} & 
        {\fontsize{50}{14}\selectfont LDC} & 
        {\fontsize{50}{14}\selectfont LPC} \\ \hline
        {\fontsize{50}{14}\selectfont Train} & 
        {\fontsize{50}{14}\selectfont 12114} & 
        {\fontsize{50}{14}\selectfont 30000} & 
        {\fontsize{50}{14}\selectfont 21500} & 
        {\fontsize{50}{14}\selectfont 22014} & 
        {\fontsize{50}{14}\selectfont 22014} & 
        {\fontsize{50}{14}\selectfont 42870} \\ \hline
        {\fontsize{50}{14}\selectfont Valid} & 
        {\fontsize{50}{14}\selectfont 1500} & 
        {\fontsize{50}{14}\selectfont 3750} & 
        {\fontsize{50}{14}\selectfont 2000} & 
        {\fontsize{50}{14}\selectfont 2750} & 
        {\fontsize{50}{14}\selectfont 2750} & 
        {\fontsize{50}{14}\selectfont 5366} \\ \hline
        {\fontsize{50}{14}\selectfont Test} & 
        {\fontsize{50}{14}\selectfont 2500} & 
        {\fontsize{50}{14}\selectfont 3750} & 
        {\fontsize{50}{14}\selectfont 1500} & 
        {\fontsize{50}{14}\selectfont 2761} & 
        {\fontsize{50}{14}\selectfont 2761} & 
        {\fontsize{50}{14}\selectfont 5357} \\ \hline
    \end{tabular}}
    \caption{Dataset Distribution}
    \label{Dataset Distribution}
\end{table}


\section{Methodolgy}
We investigated the distinctions between LLM-generated and human-generated annotations for the Indic language, Marathi, using a comparative methodology, and analyzed the results with fine-tuned BERT-based models for detailed insights.

\subsection{LLMs}
In our annotation experiments, we evaluated the performance of LLMs for the Marathi language using two prompting techniques: zero-shot and few-shot learning. We tested both open-source models (Llama 3.1 8B, Llama 3.1 405B, Gemma 2 2B, and Gemma 2 9B) and closed-source models (Gemini 1.0 Pro, GPT-4o), and compared their results with BERT-based models. The performance of each LLM under both prompting strategies is summarized in Table \ref{tab:Results}.

\subsection{BERT Based Models}

We used fine-tuned BERT-based models to compare performance with LLMs, where {\href{https://huggingface.co/l3cube-pune/marathi-sentiment-md}{MahaSent-MD}}
\href{https://huggingface.co/l3cube-pune/mahahate-bert}{MahaHate-BERT}, \href{https://huggingface.co/l3cube-pune/marathi-topic-short-doc}{MahaNews-SHC-BERT}, 
\href{https://huggingface.co/l3cube-pune/marathi-topic-medium-doc}{MahaNews-LPC-BERT}, 
and \href{https://huggingface.co/l3cube-pune/marathi-topic-long-doc}{MahaNews-LDC-BERT} are fine-tuned versions of \href{https://huggingface.co/l3cube-pune/marathi-bert-v2}{MahaBERT}, while \href{https://huggingface.co/l3cube-pune/mahahate-multi-roberta}{MahaHate-multi-RoBERTa} has \href{https://huggingface.co/l3cube-pune/marathi-roberta}{MahaRoBERTa} as the base model. Each of these models was fine-tuned on the corresponding datasets, and their respective performances are summarized in Table \ref{tab:Results}.

\section{Experimental Setup}

Our main objective is to assess the LLMs on three different tasks and related datasets to ascertain whether LLMs could take the place of, or at least support, human annotation efforts. We employed both few-shot and zero-shot prompting techniques, with LLM-generated annotations evaluated against the ground truth labels. For all datasets, the test split was used. The open-source models (Llama 3.1 8B, Gemma 2 2B, and Gemma 2 9B) exhibited slower response times and required significant computational resources to generate predictions. However, by utilizing NVIDIA NIM APIs, we were able to accelerate predictions from these models, improving both speed and precision. For the closed-source Gemini 1.0 Pro model, we used the Gemini API, while GPT-4o predictions were generated manually via ChatGPT's default settings to annotate the samples. 
In our research, we could only use a subset of samples from each dataset due to the restrictive usage regulations and cost limits of the mentioned APIs. To maintain consistency and fairness in the performance comparison, all results from both LLM-based and BERT-based models were evaluated on a uniform subset. Specifically, we evaluated 490 samples from the MahaSent and MahaHate datasets, while for MahaNews, we selected 40 samples from each of the 12 classes, amounting to a total of 480 samples.

\section{Result}
This section provides a detailed overview of the experiments conducted for the annotation of three distinct tasks, utilizing six large language models (LLMs) and six BERT-based models (BERT model fine-tuned on target task). Table \ref{tab:Results} summarizes the performance metrics of the fine-tuned BERT-based models, offering a comparative analysis against the performance of each LLM under both few-shot and zero-shot prompting scenarios. The table facilitates a comprehensive evaluation by highlighting key outcomes, enabling a thorough understanding of how each model performs across the different annotation tasks and prompting methods.

\subsection{Key Findings}
Our extensive experiments revealed crucial insights, showing that Large Language Models (LLMs) are not yet fully equipped to serve as reliable annotators for the Marathi language. The disparity between LLM-based and human-generated annotations remains substantial. Even for straightforward tasks like news classification, LLM performance was suboptimal. For more complex tasks, such as the 4-class MahaHate classification, their performance was notably disappointing, as evidenced in Table \ref{tab:Results}.

Among the LLMs evaluated, GPT-4o achieved the best results compared to others, including Llama 3.1 8B, Gemma 2 (2B and 9B), and Gemini 1.0 pro. However, both open-source and closed-source LLMs exhibited notable limitations in providing accurate and reliable annotations. Our results also demonstrate that closed LLMs like GPT-4o and Gemini 1.0 Pro outperform open LLMs namely Llama and Gemma 2B and 9B but they still underperform when compared to finetuned BERT for almost all datasets.

Our evaluation ranks the models as GPT-4o > Llama 3.1 405B > Gemini 1.0 Pro > Gemma 2 9B > Llama 3.1 8B > Gemma 2 2B, highlighting that open source Llama 3.1 405B outperforms Gemini 1.0 Pro and is second only to GPT-4o.

Compared to zero-shot prompting, few-shot prompting produced more accurate results because it gave examples of the desired input-output behavior, helping the model to understand the task's context and expectations better. We observe an absolute increase in the average accuracy of few-shot prompting by 5.3\%, 0.5\%, 3.2\%, 0.5\%, 2.2\%, and 0.8\% compared to zero-shot prompting for models Llama 3.1 8B, Gemma 2 2B, Gemma 2 9B, Gemini 1.0 Pro, Llama 3.1 405B, and GPT-4o, respectively.
While few-shot prompting techniques yielded better accuracy than zero-shot approaches, they still fell short of the performance delivered by BERT-based models.

The average accuracy gap between open-source and closed-source models is 6.7\%, while the difference between closed-source models and fine-tuned BERT-based models is 13.9\%, highlighting the lack of effective LLMs for low-resource languages.

BERT-based fine-tuned models on target task outperformed LLMs in the classification tasks for Marathi language because finetuning enabled better knowledge extraction and alignment with the task's requirements. On the other hand, LLMs, despite being trained on vast amounts of general data, lack performance on low resource languages and task-specific optimization, which limits their ability to extract the most relevant features for a specific task. This suggests that, despite the increasing popularity of LLMs, BERT-based models continue to be highly relevant, particularly for Indic languages.

We also note that the difference in the results of BERT-based models and the LLMs is comparatively less for easy tasks like the Sentiment Analysis Task, i.e. in the MahaSent dataset. At the same time, the gap is significantly higher for complex tasks like the Hate Classification and News Classification tasks in favor of BERT-based models.

\section{Conclusion}

Our study demonstrates that while LLMs like GPT, Gemini, Gemma, and Llama show potential, they currently fall short of being reliable annotators for low-resource languages like Marathi, particularly for complex tasks. BERT-based models continue to outperform LLMs in these contexts, suggesting they remain essential for accurate annotation in Indic languages. These findings indicate that further advancements are required in LLMs to make them viable alternatives for human annotations. This research highlights the need for developing more robust models tailored to the specific details of low-resource languages. This includes the creation of higher-quality, task-specific datasets for low-resource languages, ensuring better representation and reducing biases. Enhanced datasets combined with domain-specific knowledge can significantly improve annotation accuracy.
\section*{Acknowledgement}
This work was carried out under the mentorship of L3Cube, Pune. We would like to express our gratitude towards our mentor for his continuous support and encouragement. This work is a part of the L3Cube-MahaNLP project \cite{joshi2022l3cube-mahanlp}.

\bibliography{main}
\appendix
\section{Appendix}
Below are the Few-Shot prompts used for annotation tasks.

\vspace{1pt}

\begin{figure}[H]
    \centering
    \begin{minipage}{0.48\textwidth}
        \centering
        \includegraphics[width=\linewidth, scale=0.9]{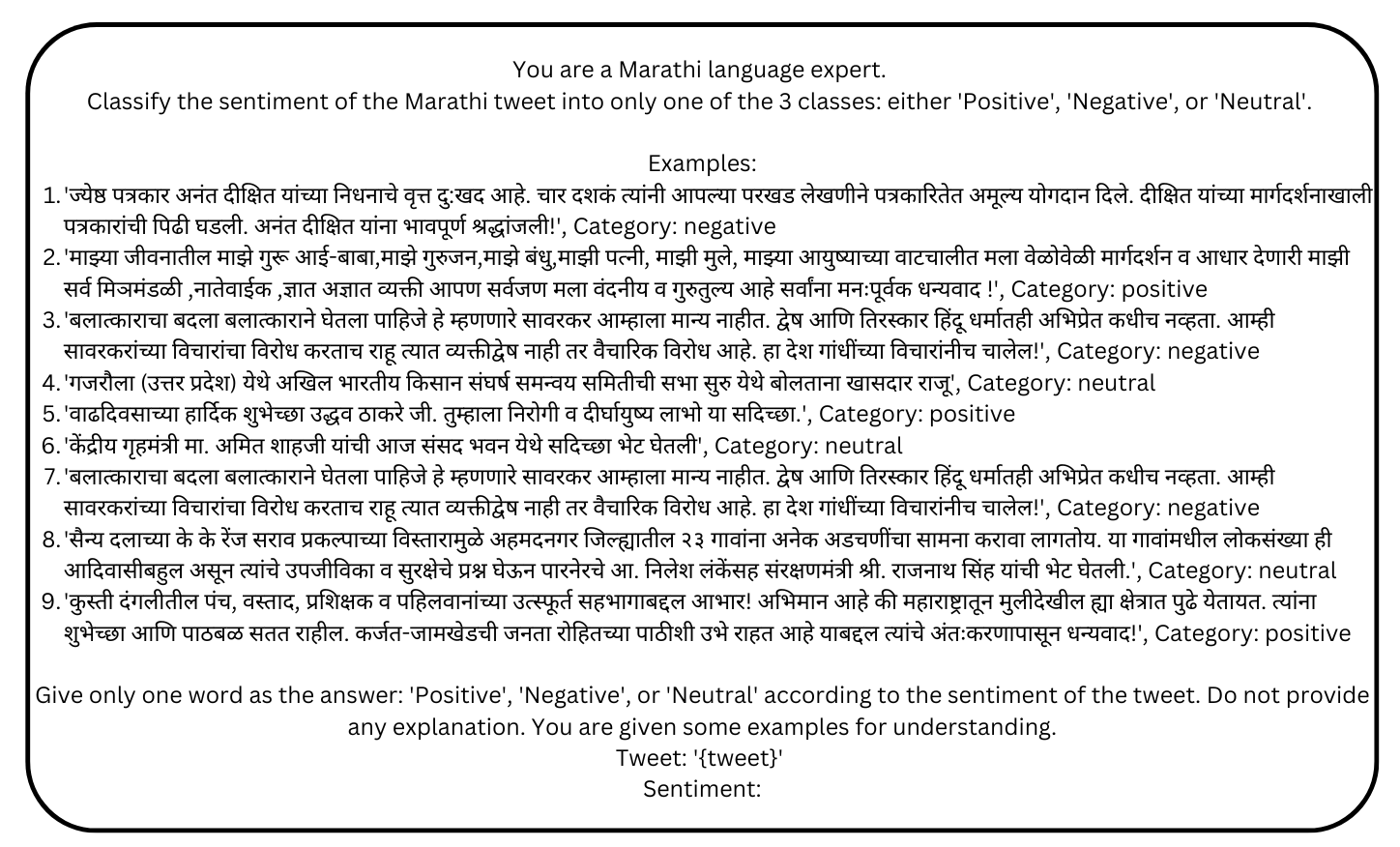}
        \caption{Few-shot prompt for MahaSent dataset.}
        \label{fig:BERT_Arch1}
    \end{minipage}
    \hfill
    \begin{minipage}{0.48\textwidth}
        \centering
        \includegraphics[width=\linewidth, scale=0.9]{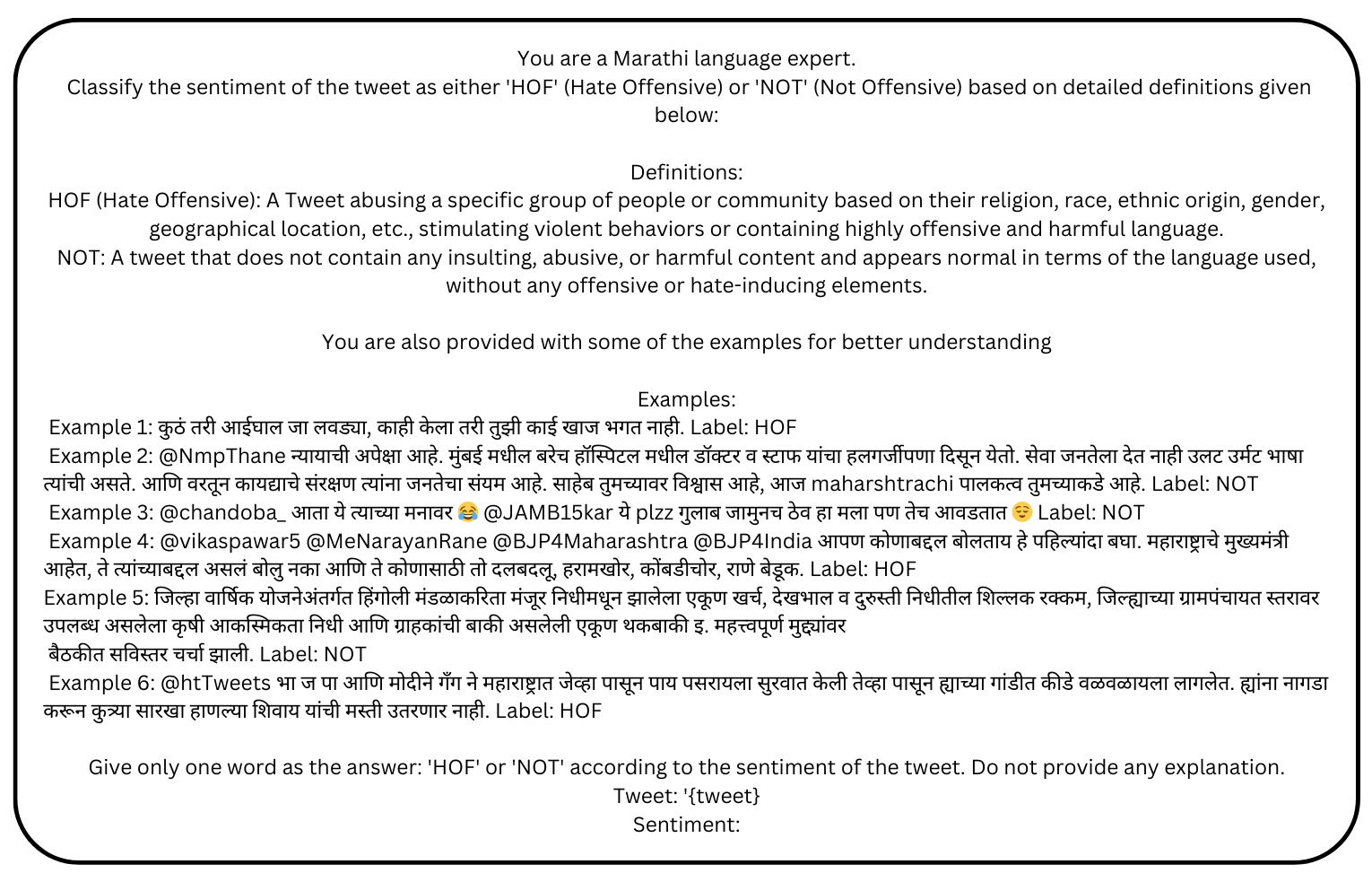}
        \caption{Few-shot prompt for MahaHate 2-Class dataset.}
        \label{fig:BERT_Arch2}
    \end{minipage}
    
    \vspace{5pt}  
    
    \begin{minipage}{0.48\textwidth}
        \centering
        \includegraphics[width=\linewidth, scale=0.9]{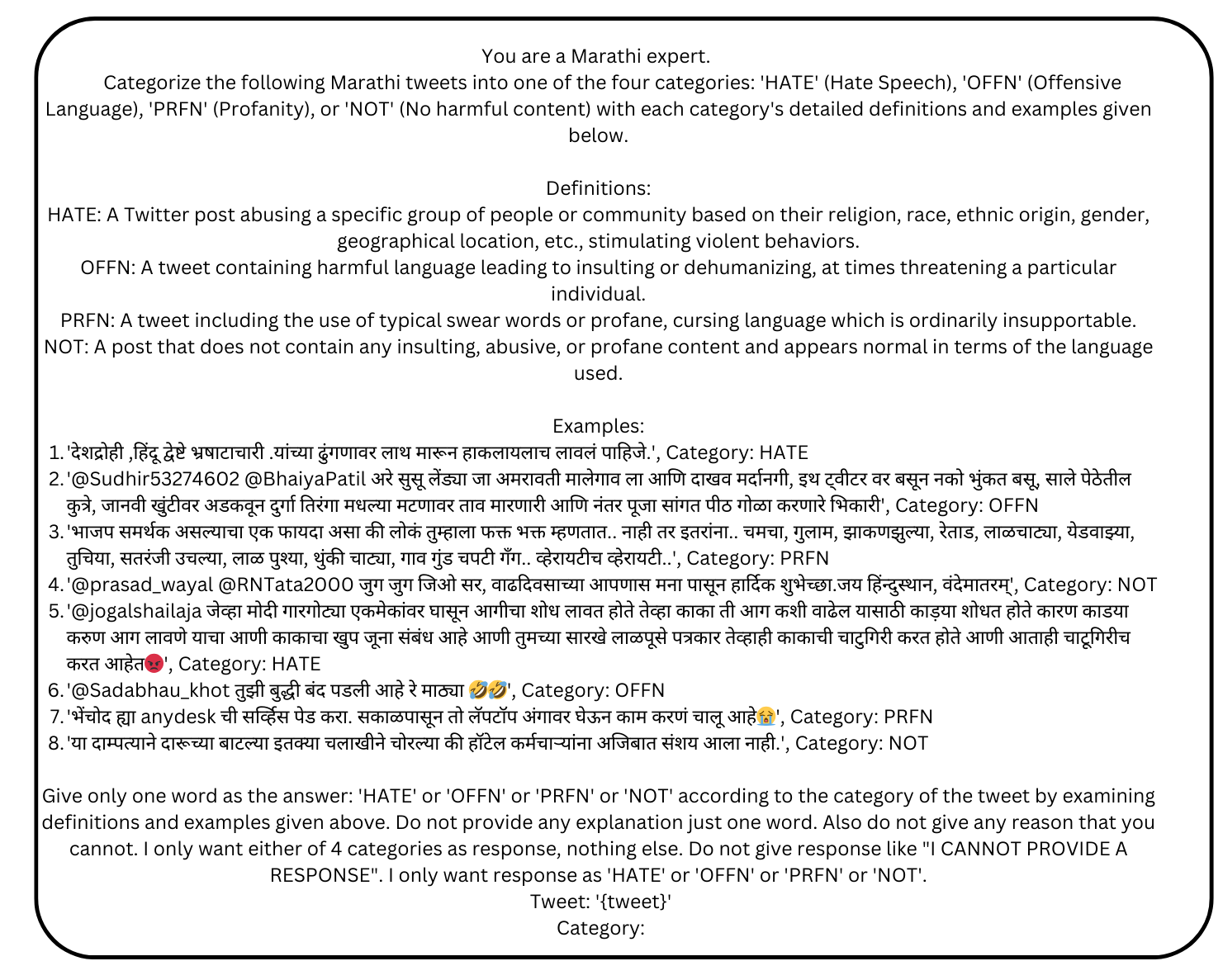}
        \caption{Few-shot prompt for MahaHate 4-Class dataset.}
        \label{fig:BERT_Arch3}
    \end{minipage}
    \hfill
    \begin{minipage}{0.48\textwidth}
        \centering
        \includegraphics[width=\linewidth, scale=0.9]{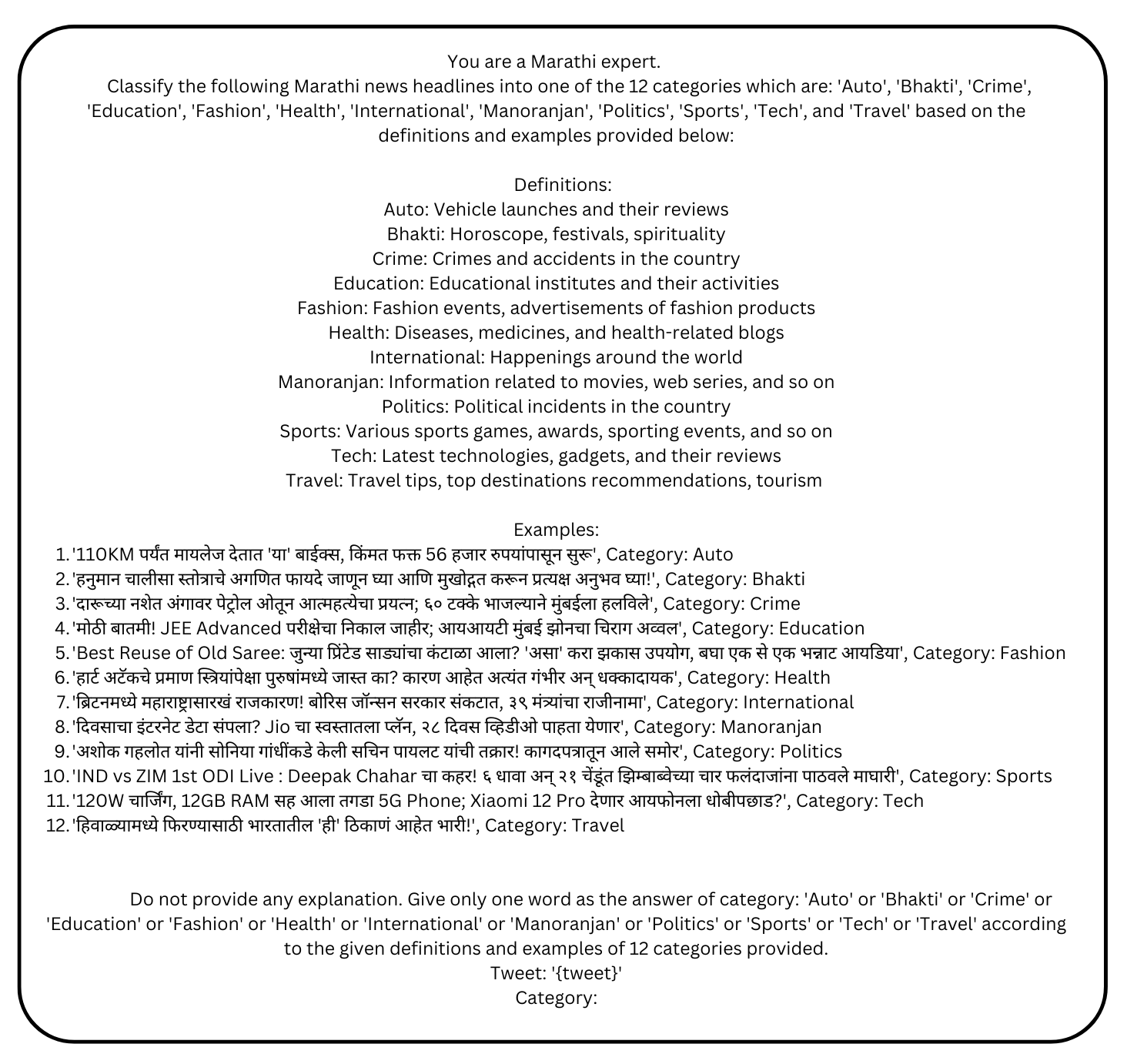}
        \caption{Few-shot prompt for MahaNews dataset.}
        \label{fig:BERT_Arch4}
    \end{minipage}
\end{figure}

\end{document}